\documentclass[11pt,a4paper]{article}
\usepackage[hyperref]{acl2021}
\usepackage{times}
\usepackage{latexsym}

\usepackage{microtype}

\aclfinalcopy 

\usepackage{amsmath}
\usepackage{booktabs}
\usepackage{multirow}
\usepackage{subcaption}
\usepackage[switch]{lineno}
\usepackage{color} 
\usepackage{graphicx} 
\usepackage[utf8]{inputenc}

\newcommand{\dataname}{{\textsc{CLIP}}}

\title{\dataname: A Dataset for Extracting Action Items for Physicians from Hospital Discharge Notes}

\author {
        James Mullenbach$^{1}$\thanks{\quad Equal contribution}\quad Yada Pruksachatkun$^{2*}$\thanks{\quad Work done while at ASAPP.}\quad Sean Adler$^{1}$ \quad Jennifer Seale$^{3}$\footnotemark[2] \\ 
        \textbf{Jordan Swartz \quad T. Greg McKelvey$^{4}$\footnotemark[2] \quad Hui Dai \quad Yi Yang$^{1}$ \quad David Sontag$^{1,5}$} \\
        $^1$ ASAPP \\
        $^2$ Amazon Alexa AI \\
        $^3$ CUNY Graduate Center \\
        $^4$ United States Digital Service \\
        $^5$ MIT
}

\begin{document}

\maketitle

\begin{abstract}
Continuity of care is crucial to ensuring positive health outcomes for patients discharged from an inpatient hospital setting, and improved information sharing can help. To share information, caregivers write discharge notes containing action items to share with patients and their future caregivers, but these action items are easily lost due to the lengthiness of the documents. In this work, we describe our creation of a dataset of clinical action items annotated over MIMIC-III, the largest publicly available dataset of real clinical notes. This dataset, which we call \dataname, is annotated by physicians and covers 718 documents representing 100K sentences. We describe the task of extracting the action items from these documents as multi-aspect extractive summarization, with each aspect representing a type of action to be taken. We evaluate several machine learning models on this task, and show that the best models exploit in-domain language model pre-training on 59K unannotated documents, and incorporate context from neighboring sentences. We also propose an approach to pre-training data selection that allows us to explore the trade-off between size and domain-specificity of pre-training datasets for this task.
\end{abstract}

\begin{table*}[t]
\centering \footnotesize
\begin{tabular}{p{0.10\textwidth}p{0.4\textwidth}p{0.4\textwidth}}

 \toprule
  \textbf{Action Type} & \textbf{Description} &  \textbf{Example} \\
  \midrule
 \textbf{Appointment} & Appointments to be made by the PCP, or monitored to ensure the patient attends them. & 
The patient requires a neurology consult at XYZ for evaluation. \\

\midrule
\textbf{Lab} & Laboratory tests that either have results pending or need to be ordered by the PCP. &
We ask that the patients’ family physician repeat these tests in 2 weeks to ensure  resolution. \\

\midrule
\textbf{Procedure} &
     Procedures that the PCP needs to either order, ensure another caregiver orders, or ensure the patient undergoes. &
 Please follow-up for EGD with GI. \\

\midrule
\textbf{Medication} & Medications that the PCP either needs to ensure that the patient is taking correctly, e.g. time-limited medications or new medications that may need dose adjustment. &
The patient was instructed to hold ASA and refrain from NSAIDs for 2 weeks. \\
\midrule

\textbf{Imaging} & Imaging studies that either have results pending or need to be ordered by the PCP. &
Superior segment of the left lower lobe: rounded density which could have been related to infection, but follow-up for resolution recommended to exclude possible malignancy \\

\midrule
\textbf{Patient Instructions} & Post-discharge instructions that are directed to the patient, so the PCP can ensure the patient understands and performs them. &
No driving until post-op visit and you are no longer taking pain medications. \\

\midrule
\textbf{Other} & Other actionable information that is important to relay to the PCP but does not fall under existing aspects (e.g. the need to closely observe the patient's diet, or fax results to another provider). &
 Since the patient has been struggling to gain weight this past year, we will monitor his nutritional status and  trend weights closely. \\
\bottomrule
\end{tabular}
\caption{Description and examples of action items. We created all examples specifically for the purpose of clarification, and they do not stem from any real patient source.}
\label{tab:example}
\end{table*}

\section{Introduction}
\label{sec:intro}

Transitioning patient care from hospitals to primary care providers (PCPs) can frequently result in medical errors \citep{Kripalani2007DeficitsIC}. When patients are discharged, they often require further actions to be taken by their PCP, who manages their long-term health, such as reviewing results for lab tests once they are available \citep{Moore2007TyingUL}. Yet PCPs often have many patients and little time to review new clinical documents related to a recent hospital stay \citep{baron}, so making this review fast, actionable, and accurate will be beneficial. 

Discharge notes are typically lengthy \cite{weis2014copy} and written as free text, so PCPs may fail to identify important pending actions, which inadvertently leads patients to poor outcomes. \citet{QualitativeInterview} found that PCPs considered the lack of a standardized follow-up section to be a key driver in missing follow-up action items. While discharge notes may include follow-up sections, they are typically aimed at the patient and not curated for PCP use. \citet{Jackson2015TimelinessOO} found that following up on pending clinical actions is critical for minimizing risk of medical error during care transitions, especially for patients with complex treatment plans. Automatic extraction of action items can make physicians more efficient by reducing the high cognitive load and time-consuming burden of using electronic health records \citep{TaiEHRusage,Sinsky,Singh2013InformationOA,Farri2013EffectsOT}. To our knowledge, there has been little previous work using machine learning to address this important clinical problem.

\paragraph{Potential impact}
Successful automatic extraction of action items can have several direct benefits. First, it can improve patient safety by fostering more comprehensive and complete care by PCPs. Second, it might make physicians more efficient at performing a comprehensive review of action items, which is critical as physicians spend an increasing amount of time interacting with electronic health record (EHR) systems \citep{TaiEHRusage,Sinsky}. Further, reviewing and synthesizing lengthy or complicated patient histories places a significant cognitive load on physicians, which has been associated with increased medical error \citep{Singh2013InformationOA, Farri2013EffectsOT}, so reducing this cognitive load is an area of opportunity. Finally, a working system might integrate with EHRs to automatically address certain action items such as scheduling appointments, thereby improving EHR usability and further reducing medical error.

\paragraph{Contributions}
We introduce a new clinical natural language processing task that accomplishes focused information extraction from intensive care unit (ICU) discharge notes by selecting sentences that contain \emph{action items} for PCPs or patients. An action item is a statement in a discharge note that explicitly or implicitly directs the reader to an action that should be taken as a result of the hospital stay described in the document. Given a discharge note, the task is to extract all action items in the note. We cast this task as a special case of multi-aspect document summarization, with each aspect representing an area of patient care to monitor or on which to take action (see examples in Table \ref{tab:example}). 

We create the first annotated dataset for this new task, \dataname, a dataset of 718 annotated notes from MIMIC-III \citep{mimiciii}, comprising over 100K annotated sentences. This will be, to our knowledge, one of the largest annotated datasets for clinical NLP, which tend to be smaller due to the expense of expert annotators.

We evaluate machine learning methods to tackle this task. Similar to prior work on multi-aspect extractive summarization, we employ sentence-level multi-label classification techniques \citep{hayashi20tacl}. Our proposed architecture consists of passing a sentence, and its neighboring sentences on its left and right, through a pre-trained BERT model~\citep{BERT} with minor modifications. 
Since there is limited annotated data but a wealth of {\em unlabeled} in-domain clinical notes, we also explore the impact of unsupervised learning on this task. We develop a method for task-targeted pre-training data selection, in which a model trained on the downstream task selects unlabeled document segments for fine-tuning a BERT model. We find that this focused pre-training is much faster than pre-training on all available data and achieves competitive results. Our results show that unsupervised pre-training of any form is critical to improving results.

Our code is available as open-source software \footnote{\url{https://github.com/asappresearch/clip}}, and our annotations are available via PhysioNet \footnote{As they are built on top of MIMIC-III, which PhysioNet maintains, access to our annotations requires the completion of an ethics course and a Data Use Agreement.}, to fully enable reproduction of our results and to provide a benchmark for evaluating future advances in clinical NLP in the context of this clinically important problem \cite{mullenbach2021}.

\section{Related Work and Datasets}

\paragraph{Clinical information extraction}
There has been a wealth of previous work on extracting information from clinical notes, much of which also follows an extractive summarization approach. For example, \citet{NLPMartin} extracts items such as patient smoking status and obesity comorbidities from discharge notes. \citet{liang-etal-2019-novel} created a hybrid system of regex-based heuristics, neural network models trained on pre-existing datasets, and models such as support vector machines for disease-specific extractive summarization. 

\citet{Semisupervised} developed a pseudo-labelling, semi-supervised approach, using intrinsic correlation between notes, to train extractive summarization models for disease-specific summaries. 
We differ from these efforts in that we do not aim to generate general-purpose or disease-specific summaries, rather we focus on extracting specific action items from discharge notes to facilitate care transfer. 

\paragraph{Clinical datasets}
Datasets and challenges on the extraction of medication, tests, and procedure mentions in clinical text \citep{i2b22009,i2b22010, MADE} have been released, but without the focus on providing actionable insight to PCPs. Additionally, multiple datasets \citep{i2b22011, i2b22012} have been introduced for detecting temporal and co-reference relations between parts of a note. While it may be useful for a model to have a good grasp of co-reference and temporal dependencies to understand what constitutes actionable information for a PCP, we choose to optimize directly for the end task, noting recent work demonstrating that modern pre-trained neural networks will identify and exploit such information as needed \citep{tenney2019bert}. Although on different tasks, we note that our dataset of 718 annotated documents is larger than recently released datasets, such as those from the n2c2 shared tasks. For comparison, 500 documents were annotated for adverse drug event extraction \cite{henry20202018}, 150 documents for family history extraction \cite{liu2018overview}, and 100 documents for clinical concept normalization \cite{henry20202019}. One of the largest annotated clinical datasets, emrQA, is built on 2,425 clinical notes \cite{pampari2018emrqa}.

\paragraph{Summarization}
Prior summarization work, which we build on, uses pre-trained transformer models to construct sentence representations that are contextualized with the {\em entire} document \citep{liu2019text, hayashi20tacl}. \citet{liu2019text} evaluate on three benchmark summarization datasets consisting of news articles.
They are shorter, with average document lengths from 400-800 words, whereas MIMIC-III discharge notes average over 1,400 words. \citet{liu2019text} evaluate with ROUGE scores standard in summarization, whereas we take advantage of having  ground truth extracted sentences and evaluate with classification metrics, providing a substantially different task.

\citet{liang-etal-2019-novel} develop a disease-specific summary dataset, but it is not public and their methods involve combining a mix of outputs from models performing auxiliary tasks such as concept recognition, adverse drug event extraction, and medication change detection, each of which have to be individually developed and maintained. 

\section{\dataname~Dataset}
In this section, we describe the process of creating our \dataname~dataset, short for \textsc{CLInicalfollow-uP}, and report statistics on the dataset.

\subsection{Data collection}
\label{sec:data}
\dataname~is created on top of the popular clinical dataset MIMIC-III~\citep{mimiciii}. The MIMIC-III dataset contains 59,652 critical care discharge notes from the Beth Israel Deaconess Medical Center over the period of 2001 to 2012, among millions of other notes and structured data. We annotated 718 randomly sampled discharge notes from the set of patients that were discharged from the ICU (i.e., survived) and thus brought back to the care of their primary care physician or relevant specialists. Though this dataset is orders of magnitude smaller than general summarization datasets such as \citet{nallapati2016abstractive}, we note the relatively large expense associated with clinical annotation due to both the length of documents ($\sim$160 sentences on average) and the requirement of domain experts. This dataset is also the first of its kind in the clinical space. The total number of sentences is 107,494, of which 12,079 have at least one label. 

The sampled MIMIC-III data is further split randomly into training, validation, and test sets, such that all sentences from a document go to the same set, with 518, 100, and 100 notes respectively. 

Our dataset was annotated by four physicians and one resident over the course of three months. We underwent several rounds of initial annotations with calibration processes and instruction refinement in between. Additional annotation details are provided in the appendix and the full guidelines are available on our public repository. We estimated inter-rater reliability by having two physician annotators independently annotate a set of 13 documents comprising 2600 sentences. Comparing predictions on a binary reduction of the task, in which a match indicates that both annotators labeled a sentence (regardless of chosen label types), we measured a Cohen's kappa statistic of 0.925. 

The seven action item aspects that we labeled in the dataset, along with example discharge note snippets for each one, are presented in Table \ref{tab:example}.

To emphasize the subtlety and complexity of this task, we highlight here some example rules that state what should \emph{not} be annotated. For the appointment label, we should exclude sentences that refer to ``as needed'' appointments, e.g. ``See your endocrinologist as needed.''; this describes no deviation from status quo behavior and thus does not warrant follow-up action. For the medication label, we specifically exclude sentences describing simple additions to the medication list, e.g. ``Discharged on glargine 10u at bedtime,'' as these typically do not require further action. However we \emph{include} instructions to hold and restart medications, new medications with an end date (e.g. antibiotics), and medications requiring dosage adjustment (e.g. ``...the plan is to keep patient off diuretics with monitoring of his labs and reinstitution once the kidney function improves''), as these are likely to require PCP action.

\subsection{Training set statistics}

\begin{table}[t]
\centering
\begin{tabular}{@{}lr@{}}
\toprule
Patient Instructions & 6.55\% \\
Appointments         & 4.59\% \\
Medications          & 1.88\% \\
Lab tests            & 0.69\% \\
Procedures           & 0.28\% \\
Imaging              & 0.18\% \\
Other                & 0.05\% \\ \bottomrule
\end{tabular}
\caption{Prevalence of each label type in \dataname~training set.}
\label{tab:label_prevalence}
\end{table}

Due to the large amount of discharge note text that has information not directly actionable for follow-up, most sentences remain without a label after the annotation process; 11.2\% of training set sentences have a label. Of the sentences with labels, 28.6\% have multiple labels. Table \ref{tab:label_prevalence} shows the frequency of each label type at the sentence level in the training set.

\begin{figure*}[t]
  \centering 
\begin{tabular}{cc}
        \includegraphics[scale=0.7]{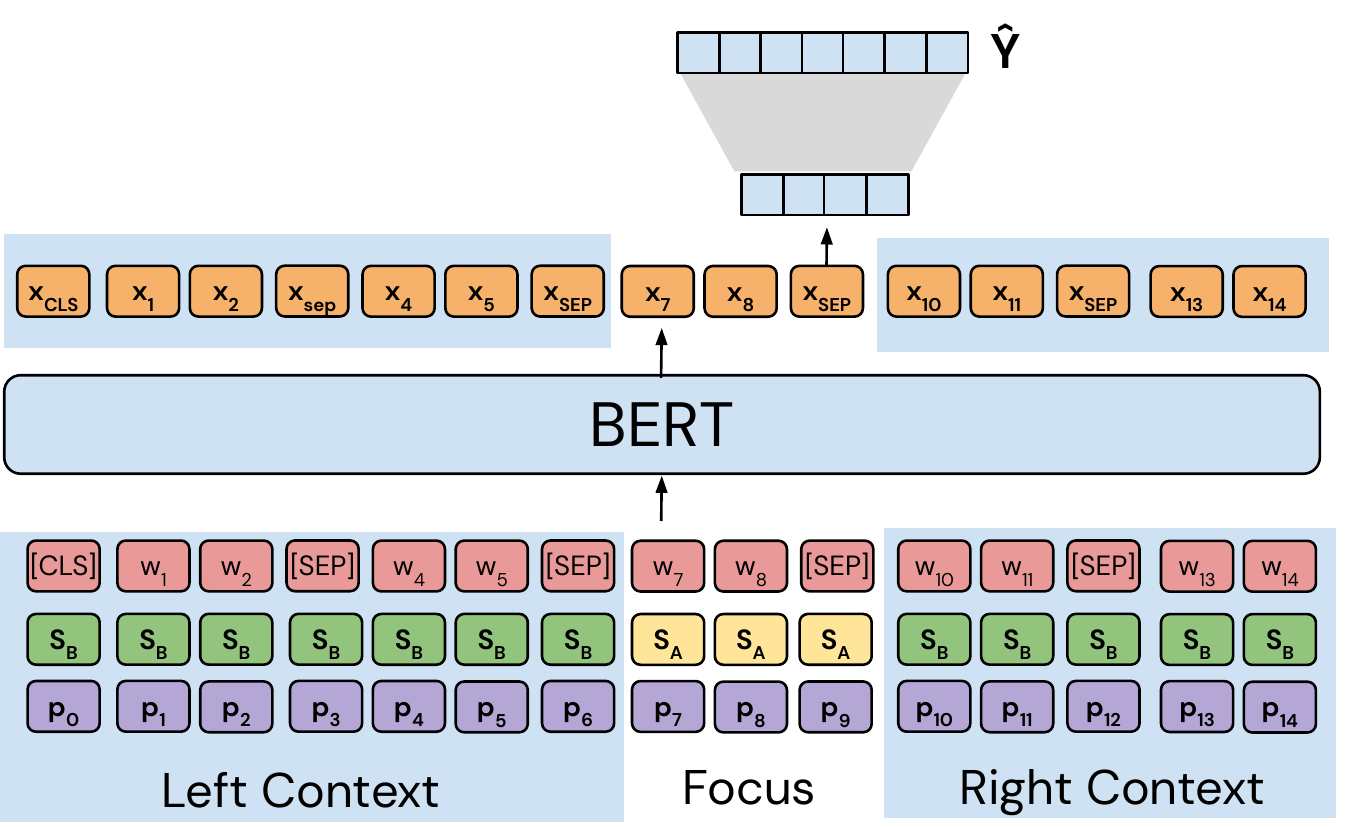} & 
    \includegraphics[scale=0.7]{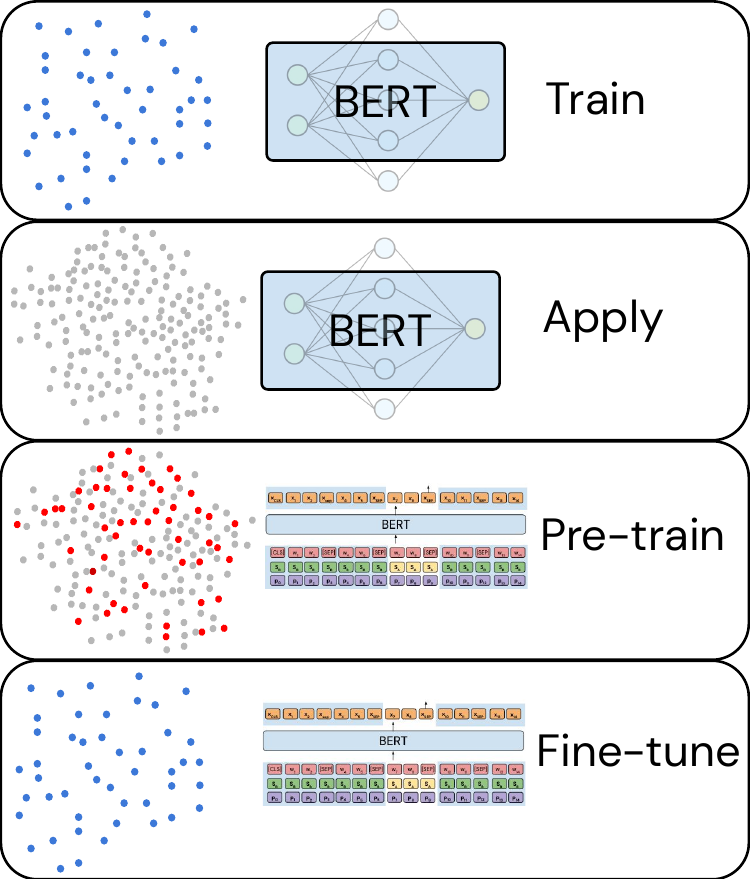} \\
    (a) & (b) \\
  \end{tabular}
  \caption{(a) Illustration of our BERT-based architecture. We input the sentence (red; top) to classify along with 2 sentences of context on either side, joined with \texttt{[SEP]} tokens and accompanied with segment (green; middle) and position (purple; bottom) embeddings, to integrate neighboring intra-document context into the token representations of the focus sentence. We then apply a linear classification layer over the \texttt{[SEP]} token representation at the end of the focus sentence. (b) Our pre-training method. First, we train a supervised model with the labeled data (blue). Then, we apply it to unlabeled data (gray) to surface a fraction of the data to pre-train the model (a) with (red). After pre-training, we fine-tune on the labeled data, which leads to similar results as pre-training with all unlabeled data. }
  \label{fig:model}
\end{figure*}

\subsection{Dataset comparison and phenomena analysis}

To distinguish the contribution of our dataset in the context of existing text summarization datasets, we performed a manual quantitative comparison between \dataname~and the summarization datasets CNN \cite{hermannteaching} and WikiASP \cite{hayashi20tacl}. For WikiASP, we chose sentences from the ``Event'' genre of summary, as our dataset describes hospital stays which could be considered events. Inspired by \citet{suhr2017corpus}, we identified five phenomena to compare across datasets - quantification (in the numerical sense, as in ``300 mg'' or ``twenty-three people''), temporal expressions, conditional expressions, imperative mood or second-person statements, and out of vocabulary (OOV) terms \footnote{By OOV, we mean any token that must be split into multiple WordPiece tokens given the vanilla BERT vocabulary. For example, the common abbreviation for ``patient'', ``pt'', becomes ``p'', ``\#\#t''.}. We gathered 100 sentences from the summaries of each dataset and counted the occurrences of each phenomena. 

We see that \dataname~has a relative wealth of imperative and second-person statements, which is not surprising due to the prevalence of patient-directed language in ``Patient instructions''-labeled sentences. \dataname~and WikiASP both have more temporal expressions than CNN, which are contained in around half of the sample sentences of each. Despite the prevalence of clinical jargon in \dataname, WikiASP actually contained the most OOV words, perhaps due to the diversity of sources of that dataset. Conditional language, such as ``If you miss any doses of this medication, your stents could clot off again...'', were uncommon in all datasets but occurred most in \dataname. 

\begin{table}[]
\resizebox{\columnwidth}{!}{%
\begin{tabular}{@{}llll@{}}
\toprule
 & CNN & WikiASP & CLIP \\ \midrule
Quantification & 27 & 26 & 27 \\
Temporal & 34 & 56 & 48 \\
Conditional & 0 & 3 & 10 \\
Imperative / 2nd person & 3 & 4 & 41 \\ 
\# OOV & 0.91 & 2.15 & 2.10 \\ \bottomrule
\end{tabular}}
\caption{Comparing sentences in existing summarization datasets with ours. We randomly sampled 100 sentences from extractive summaries in each dataset and counted each phenomenon. \# OOV is reported as the average number of OOV terms in a sentence. For CNN and WikiASP, we adopted the greedy approach of \citet{nallapati2016abstractive} to create extractive summaries.}
\label{tab:dataset_comparison}
\end{table}

\section{Learning to Extract Action Items}

With a discharge note as input, the task is to output the clinically actionable follow-up items found within the note. There are many summarization methods that could appropriately handle this problem. The length of these documents and the high relative risk of missing information in a clinical setting discourages the option of truncating documents to fit into modern neural network models which may have maximum length requirements, so we develop methods that approach the task as multi-label sentence classification. Summarization of a full document can then be accomplished with the resulting model by feeding each sentence into the model in sequence and aggregating the sentences that the model labels. We will evaluate our experiments on this multilabel classification formulation, as well as on a binary reduction of the problem in which the objective is to simply identify which sentences have any type of label. This binary framing is still useful, as surfacing the sentences for a reader is the primary objective that will save time and effort, with classification of the sentence being a secondary benefit.

\subsection{Model architecture}
The BERT architecture \citep{BERT} has been widely used within clinical NLP in the past year with successful results~\citep{BioBERT, alsentzer-etal-2019-publicly, BERTPhenotyping, johnson2020deidentification, mcdermott2020, zhang2020hurtful}. In particular, \citet{si2020students} has shown the effectiveness of BERT for use on small annotated clinical datasets, such as the one we develop. We use BERT as the basis for our proposed model.

\paragraph{BERT-based baselines}
To demonstrate baseline BERT performance, we fine-tune pre-trained BERT models on our task. As the simplest approach, we feed a sentence into BERT, take the hidden state of the [CLS] token as the sentence-level representation, and train a linear layer over that representation.
To adapt BERT to our domain, we also experiment with a previously released version of BERT which has been further pre-trained on MIMIC-III discharge notes \citep{alsentzer-etal-2019-publicly}, and fine-tune it on our task in the same way. We refer to this variant as \textsc{MIMIC-DNote-BERT}. \citet{alsentzer-etal-2019-publicly} also release a version pre-trained on all MIMIC-III notes, which we refer to as \textsc{MIMIC-Full-BERT}. Both MIMIC-Full-BERT and MIMIC-DNote-BERT are initialized with BioBERT \citep{BioBERT}, which is pre-trained on a corpus of biomedical research articles.

\paragraph{Incorporating neighboring context}
Surrounding contexts are critical for the task, for two reasons: 1) an individual sentence may not have the full picture on the type of the action; 2) neighboring sentences tend to share the same label (occurs for 27\% of sentences). So, we incorporate context beyond an individual sentence into our BERT-based sentence representations, by concatenating the two sentences each that immediately precede and follow the sentence to the input. To do this, we follow the encoder architecture of \citet{liu2019text}, which concatenates sentences with special tokens and applies alternating segment embeddings to alternating sentences. We make the following modifications: we exclude the additional transformer layers on top of the BERT output, use only \texttt{SEP} tokens to separate sentences, and apply the segment embedding $S_A$ to the tokens in the focus sentence and $S_B$ to all other tokens, as pictured in \autoref{fig:model}. We initialize models of this architecture with various pre-trained BERT parameters in experiments.

\begin{table*}
\centering
\resizebox{\textwidth}{!}{%
\begin{tabular}{@{}lll|ll|l@{}}
\toprule
\textbf{} & \multicolumn{2}{c|}{\textbf{Micro}} & \multicolumn{2}{c|}{\textbf{Macro}} & \multicolumn{1}{c}{\textbf{Binary}} \\
\textbf{Model}  & \textbf{F1} & \textbf{AUC} &  \textbf{F1} & \textbf{AUC} &  \textbf{F1} \\ \midrule
Bag-of-words+TFIDF & 0.709 & 0.958 & 0.512 & 0.937 & 0.783 \\
CNN & 0.723 (0.010) & 0.964 (0.003) & 0.540 (0.013) & 0.962 (0.002) & 0.810 (0.008) \\
BERT & 0.758 (0.008) & 0.962 (0.006) & 0.593 (0.028) & 0.963 (0.003) & 0.827 (0.006) \\
MIMIC-Full-BERT & 0.765 (0.005) & 0.971 (0.002) & 0.624 (0.016) & 0.966 (0.003) & 0.832 (0.004) \\
MIMIC-DNote-BERT & 0.767 (0.004) & 0.972 (0.006) & 0.631 (0.018) & 0.967 (0.002) & 0.834 (0.004) \\  \midrule
BERT+Context & 0.791 (0.007) & 0.947 (0.011) & 0.635 (0.013) & 0.971 (0.003) & 0.856 (0.007) \\ 
MIMIC-Full-BERT+Context & 0.794 (0.008) & 0.954 (0.010) & 0.641 (0.031) & 0.972 (0.003) & 0.857 (0.003) \\ 
MIMIC-DNote-BERT+Context & 0.796 (0.012) & 0.958 (0.015) & 0.661 (0.025) & 0.977 (0.003) & 0.856 (0.008) \\
TTP-BERT+Context (2M) & 0.809 (0.006) & 0.957 (0.008) & 0.660 (0.013) & 0.973 (0.004) & 0.865 (0.003) \\
TTP-BERT+Context (1M) & 0.802 (0.004) & 0.959 (0.010) & 0.654 (0.012) & 0.974 (0.003) & 0.857 (0.006) \\
TTP-BERT+Context (500k) & 0.803 (0.005) & 0.953 (0.016) & 0.671 (0.017) & 0.976 (0.004) & 0.862 (0.005) \\
TTP-BERT+Context (250k) & 0.807 (0.009) & 0.962 (0.010) & 0.668 (0.028) & 0.975 (0.002) & 0.866 (0.007) \\ \bottomrule
\end{tabular}}
\caption{Experiment results on the \dataname~test set. We report results as an average of at least 10 runs with varying random seeds, with standard deviations in parentheses. Models using context sentences are listed in decreasing order of amount of pre-training data used.}
\label{tab:main_results}
\end{table*}

\subsection{Task-targeted pre-training}

Given the limited amount of annotated data, we are motivated to pursue semi-supervised approaches. We seek to explore the trade-off between generalized and domain- or task-specific data for language model pre-training, by introducing a technique for targeted pre-training which we call \textbf{T}ask-\textbf{T}argeted \textbf{P}re-training (TTP). TTP requires less data and computation, yet attains comparable performance to pre-training on large in-domain datasets that prior work studied \citep{alsentzer-etal-2019-publicly}. The goal of this approach is to surface unlabeled sentences that may be positive examples, in the vein of self-supervision techniques such as Snorkel \cite{ratner2017snorkel}. In contrast to Snorkel, which uses model predictions to generate pseudolabels to train with, TTP uses model predictions to select sentences for {\em pre}-training, using auxiliary tasks.

To create a task-targeted dataset, we first fine-tune a vanilla BERT model on our task, and then we use the learned model to classify all unlabeled sentences. We select all sentences that the model predicts as having action items, using a fixed threshold. Due to the multi-label nature of our task, we apply the threshold across all labels and select sentences in which at least 1 label score is above the threshold. The threshold used to select the task-targeted sentences can be tweaked to create datasets for pre-training that are smaller and more task-focused (for higher thresholds), or larger and more general (for lower thresholds), which we experiment with. This approach is inspired by and similar to task-adaptive pre-training (TAPT) introduced by \citet{gururangan2020don}. In that work, a pre-trained bag-of-words language model encodes sentences in labeled and unlabeled datasets, and for each labeled sentence selects its nearest neighbor unlabeled sentences according to the model. In this paper, we select data points using the full prediction model (rather than just an encoder), and use thresholding which provides maximal control over the size of the selected dataset. Further, directly applying TAPT to our case may not work well as it does not distinguish positive and negative samples in the in-domain dataset, so the surfaced sentences from TAPT may be less relevant. Our approach benefits from using an encoding method that is trained on the task we are targeting. 

After selecting data, we pre-train a BERT-Context model on the targeted dataset, pulling in neighboring sentences of the targeted sentences. As auxiliary tasks, we used masked language modeling (MLM) and a sentence switching task \cite{wang2019self}. For MLM, we mask tokens in the context sentence only, independently with probability 0.15. For sentence switching, with probability 0.25 we swap the focus sentence with another randomly chosen sentence from the same document, and predict whether the focus sentence was swapped using the context sentences. Cross entropy losses for both tasks are computed and summed to compute the total loss for an instance. These tasks encourage the model to learn how to incorporate information from the context sentences into its representation. \autoref{fig:model} depicts the entire process. This process can be repeated, by using the final resulting model to then select a new set of sentences for pre-training, however we did not experiment with this as one iteration was enough to produce competitive results.

\section{Evaluation}

\subsection{Data preparation and model training}

We first generate synthetic surrogates for entities redacted during de-identification, apply a custom sentence tokenizer adapted from open-source software \footnote{https://github.com/fnl/syntok} \footnote{https://github.com/wboag/mimic-tokenize}
to tokenize the document into sentences, and  lower case every sentence.
Discharge notes in MIMIC often have semi-structured sections, with headers denoting them, e.g. \texttt{BRIEF HOSPITAL COURSE:}, which the tokenizer is built to identify.

Using TTP, we select pre-training datasets of sizes $\sim$250K, $\sim$500K, $\sim$1M, and $\sim$2M sentences from the set of MIMIC-III discharge notes.

As baselines, we train a TF-IDF-weighted bag-of-words logistic regression model with L1 regularization and a max-pooling 1-D convolutional neural network (CNN). The CNN is initialized with BioWordVec vectors \citep{biowordvec1, biowordvec2}, which are trained on PubMed and MIMIC-III notes, and the CNN is trained with the binary cross-entropy (BCE) loss.

All BERT-based models are loaded, pre-trained as appropriate, and fine-tuned using the transformers library \citep{Wolf2019HuggingFacesTS}, using BCE loss, and backpropagating and applying gradient updates through all of BERT's parameters. We used library default parameters, except for the batch size which we adjusted to 32 based on validation set performance and training stability. All neural models are trained with early stopping on the macro-averaged AUROC metric. Early stopping is also applied to the pre-training step, using the loss on an unlabeled held-out set as the criterion. 
\begin{table*}
\centering
\resizebox{\textwidth}{!}{%
\begin{tabular}{@{}lrrrrrrr@{}}
\toprule
\textbf{Model} & \multicolumn{1}{l}{\textbf{Patient}} & \multicolumn{1}{l}{\textbf{Appt}} & \multicolumn{1}{l}{\textbf{Medication}} & \multicolumn{1}{l}{\textbf{Lab}} & \multicolumn{1}{l}{\textbf{Procedure}} & \textbf{Imaging} & \multicolumn{1}{l}{\textbf{Other}} \\ \midrule
Bag-of-words & 0.741 & 0.792 & 0.546 & 0.625 & 0.302 & 0.343 & 0.236 \\
CNN & 0.759 & 0.824 & 0.595 & 0.629 & 0.315 & 0.431 & 0.228 \\
BERT & 0.780 & 0.855 & 0.635 & 0.719 & 0.415 & 0.474 & 0.275 \\
MIMIC-DNote-BERT & 0.783 & 0.854 & 0.656 & 0.741 & 0.524 & 0.567 & 0.294 \\ \midrule
MIMIC-DNote-BERT+Context & 0.830 & 0.882 & 0.659 & 0.744 & 0.597 & 0.567 & 0.349 \\
TTP-BERT+Context (250k) & 0.841 & 0.887 & 0.668 & 0.745 & 0.548 & 0.566 & 0.365 \\
\end{tabular}}
\caption{Average balanced F1 scores on the test set for each label across 10 runs.}
\label{tab:per_label_results}
\end{table*}

\subsection{Reported metrics}

We report results on the test set using micro- and macro-averaged metrics common in multilabel classification, and F1 for the binary reduction of the task. Micro-averaged metrics treat each (sentence, label) pair as an individual binary prediction, and macro-averaged metrics compute the metric per-label and then average these results across labels. For binary F1, we transform the label and model predictions into binary variables indicating whether any type of label was predicted for the sentence, and then calculate metrics, ignoring whether the types of the predicted labels were accurate. 

\subsection{Choosing prediction thresholds}

To ensure the fairest comparison between models and eliminate some arbitrariness in results that may arise when training on imbalanced data and evaluating with a fixed 0.5 threshold, we also tune thresholds for each label such that its F1 score on the validation set is maximized. For micro metrics, we choose the threshold that provides the highest micro F1 score. We then apply these thresholds when evaluating on the test set.

\section{Results}

The main set of results are reported in \autoref{tab:main_results}.
Models pre-trained with TTP have the size of their pre-training dataset denoted in parentheses. BERT and both MIMIC-BERT models outperform the logistic regression and CNN baselines. The results using MIMIC-DNote-BERT demonstrate the importance of domain-specific pre-training; it improves in all metrics over BERT. Using neighboring sentences, as we do in ``-Context'' models, also provides a performance boost across all metrics save for Macro AUC, comparing MIMIC-DNote-BERT to MIMIC-DNote-BERT+Context. To compare with human performance, our inter-annotator agreement on the binary task, measured in terms of F-1, was 0.930, and the highest mean binary-F1 from the model evaluations approaches 0.86.

When using just 250,000 sentences from the MIMIC discharge notes for pre-training (TTP-BERT-Context 250K), task results are competitive with and in some cases exceed MIMIC-DNote-BERT+Context, which is pre-trained on all MIMIC discharge notes, which contain ~9M sentences. Our TTP approach is able to complete domain-specific pre-training within $\sim$12 hours, while \citet{alsentzer-etal-2019-publicly} report a pre-training time of 17-18 days for MIMIC-Full-BERT. 

We next investigate results on each label (see \autoref{tab:per_label_results}), for a subset of models. The in-domain pre-training for MIMIC-DNote-BERT models provides gains for nearly all label types, and including context also gives a boost to the F1 score of most labels. All models perform poorly predicting the ``Other'' label, which encompasses a long tail of many different types of follow-ups which we did not further categorize, making modeling difficult. 
Imaging and Procedure label performance lags others, likely due to their lower prevalence (\autoref{tab:label_prevalence}).

\subsection{Error analysis}

We examine errors made by TTP-BERT-Context (1M), focusing on false negatives, the most costly type of error in this use case. Inspection of the test set with physician input yields two high-level phenomena of the data that occur repeatedly in error cases: clinical jargon / knowledge, and temporal expressions / conditional language.

\paragraph{Clinical jargon}

Perhaps the most obvious drawback of applying general-purpose language models to clinical language data is that clinical language is heavily laden with clinical jargon, abbreviations, and misspellings. Although the WordPiece tokenization used by BERT-based models can tokenize any input, the more separation of clinical terms happens, the more model capacity is reduced, as lower layers in BERT have to learn how to combine the meaning of the WordPieces into word-level representations. We observed several cases in which even common clinical jargon may have interfered with the model's performance in an otherwise unambiguous sentence. Bolded words are OOV's:  \texttt{<-please take medications as directed -follow up with \textbf{pcp} mark carter using>}, \texttt{<plan for repeat chest \textbf{xray} pa/\textbf{lat} and \textbf{lordotic} view to \textbf{reevaluate} when returns 12-18 for wound check>}.

Many cases of this type of error also suggest that a lack of explicit clinical knowledge could be a barrier, in addition to the technical issue of WordPiece tokenization. In this example \texttt{promethazine} is a drug that can be prescribed for a short defined period: \texttt{3 . promethazine 25 mg tablet sig : 0.5 tablet po q6h ( every 6 hours ) as needed for nausea .} In the following example, the procedures described are required but do not need an appointment, and the model erroneously applied the Appointment label:  \texttt{however , the patient will need aggressive pulmonary toilet including good oral suctioning care and chest pt as pt is at risk for aspiration .}

\paragraph{Temporal expressions}

The model may also struggle with temporal expressions, which are common especially in the ``Medication'' label type. This label is intended to surface cases of medications that need to be tweaked, started, or stopped after a specified time period. Example: \texttt{...you should go back to your regular home dosing of 20 units in the morning and 24 units at dinner time after completing your prednisone }. While many training examples gave explicit durations (e.g. ``for 14 days''), many of the false negative examples described dependencies between future patient actions, including with conditional ``if'' statements. Example: \texttt{if he needs further management he may do well with clonidine .}
\section{Discussion}

Our results show that the common regime of fine-tuning a large pre-trained model is a useful method for our task of extracting clinical action items. Additionally, we investigated the trade-off between task-specificity and pre-training data size, and found our task-targeted pre-training method enables one to navigate this trade-off, producing models with comparable performance on the end task that require less data for pre-training. While trading off these concerns may not be needed if effective public models exist for a given task, we believe this technique is useful in scenarios in which users have large, domain-specific, private datasets and specific tasks in mind. This is often the case for healthcare institutions and developers of clinical machine learning software, as privacy concerns tend to preclude data sharing between institutions.

From a modeling perspective, there are many possible avenues for future work. Taking a structured prediction lens and leveraging sentence-level label dependencies or applying structured prediction models could be helpful, although \citet{cohan2019pretrained} note that CRF layers did not improve their performance for a sequential sentence classification task. 
We acknowledge that our sentence classification approach is a simplification of the more general span detection problem, and this approach could bring improved precision by focusing on which parts of sentences matter, which may be important as we found that sentence tokenization was non-trivial for clinical notes.

Finally, the question of whether such an approach to follow-up workflow augmentation is successful in increasing patient safety, clinician efficiency, or EHR usability is an empirical one. We hope to evaluate in the future whether a highlighted note such as one these models could provide will reduce the time a physician takes to, for example, answer certain questions about a patient's hospital stay.
In alignment with recent calls for increased rigor in the evaluation of machine learning-derived clinical decision support systems \citep{Kelly2019KeyCF}, future work should include further prospective, controlled evaluation of the generalizability, stability, interpretability, unbiasedness, usability, and efficacy of this approach. We hope that our dataset and initial model development can lay the groundwork for future investigation.

\section{Conclusion} 

 We introduce the task of detecting clinical action items from discharge notes to help primary care physicians more quickly and comprehensively identify actionable information, and present the \dataname~
dataset, which we will release to the community. Given perfect performance, this would reduce the number of sentences a PCP may need to read by 88\%. The best model's binary F1 is near 0.9, compared to the human benchmark of 0.93. 
These models could additionally be used for clinical research. For example, a calibrated model could derive statistics for how often each type of action item is seen for different patient populations, which can provide insight into typical patient or PCP burden after hospital discharge. 

We evaluated BERT-based models that incorporate multi-sentence context, and introduced a novel task-targeted pre-training approach that can reduce pre-training time while maintaining similar performance to models pre-trained on much larger in-domain datasets. The models have promising results, however we anticipate there is still room for improvement, particularly for the rare labels. 

We encourage the clinical NLP community to further investigate the problem of detecting action items from hospital discharge notes, which can help improve reliably safe transitions of care for the most vulnerable patients. 

\section*{Acknowledgements}

We thank our team of physician annotators for their fruitful collaboration and the reviewers for their comments which improved this paper.

\bibliographystyle{acl_natbib}
\bibliography{acl2021}

\begin{thebibliography}{46}
\expandafter\ifx\csname natexlab\endcsname\relax\def\natexlab#1{#1}\fi

\bibitem[{Alsentzer et~al.(2019)Alsentzer, Murphy, Boag, Weng, Jindi, Naumann,
  and McDermott}]{alsentzer-etal-2019-publicly}
Emily Alsentzer, John Murphy, William Boag, Wei-Hung Weng, Di~Jindi, Tristan
  Naumann, and Matthew McDermott. 2019.
\newblock \href {https://doi.org/10.18653/v1/W19-1909} {Publicly available
  clinical {BERT} embeddings}.
\newblock In \emph{Proceedings of the 2nd Clinical Natural Language Processing
  Workshop}, pages 72--78, Minneapolis, Minnesota, USA. Association for
  Computational Linguistics.

\bibitem[{Baron(2010)}]{baron}
Richard~J. Baron. 2010.
\newblock What's keeping us so busy in primary care? a snapshot from one
  practice.
\newblock \emph{The New England Journal of Medicine}, 362 17:1632--6.

\bibitem[{{Chen} et~al.(2019){Chen}, {Peng}, and {Lu}}]{biowordvec2}
Q.~{Chen}, Y.~{Peng}, and Z.~{Lu}. 2019.
\newblock Biosentvec: Creating sentence embeddings for biomedical texts.
\newblock In \emph{Proceedings of the 2019 IEEE International Conference on
  Healthcare Informatics (ICHI)}, pages 1--5.

\bibitem[{Cohan et~al.(2019)Cohan, Beltagy, King, Dalvi, and
  Weld}]{cohan2019pretrained}
Arman Cohan, Iz~Beltagy, Daniel King, Bhavana Dalvi, and Daniel~S Weld. 2019.
\newblock Pretrained language models for sequential sentence classification.
\newblock In \emph{Proceedings of the 2019 Conference on Empirical Methods in
  Natural Language Processing and the 9th International Joint Conference on
  Natural Language Processing (EMNLP-IJCNLP)}, pages 3684--3690.

\bibitem[{Devlin et~al.(2019)Devlin, Chang, Lee, and Toutanova}]{BERT}
Jacob Devlin, Ming-Wei Chang, Kenton Lee, and Kristina Toutanova. 2019.
\newblock \href {https://aclweb.org/anthology/papers/N/N19/N19-1423/} {Bert:
  Pre-training of deep bidirectional transformers for language understanding}.
\newblock In \emph{Proceedings of the 2019 North American Chapter of the
  Association for Computational Linguistics: Human Language Technologies
  (NAACL-HLT)}, pages 4171--4186.

\bibitem[{Farri et~al.(2013)Farri, Monsen, Pakhomov, Pieczkiewicz, Speedie, and
  Melton}]{Farri2013EffectsOT}
Oladimeji Farri, Karen~A. Monsen, Serguei V.~S. Pakhomov, David~S.
  Pieczkiewicz, Stuart~M. Speedie, and Genevieve~B. Melton. 2013.
\newblock Effects of time constraints on clinician-computer interaction: A
  study on information synthesis from ehr clinical notes.
\newblock \emph{Journal of Biomedical Informatics}, 46 6:1136--44.

\bibitem[{Gururangan et~al.(2020)Gururangan, Marasovi{\'c}, Swayamdipta, Lo,
  Beltagy, Downey, and Smith}]{gururangan2020don}
Suchin Gururangan, Ana Marasovi{\'c}, Swabha Swayamdipta, Kyle Lo, Iz~Beltagy,
  Doug Downey, and Noah~A Smith. 2020.
\newblock Don’t stop pretraining: Adapt language models to domains and tasks.
\newblock In \emph{Proceedings of the 58th Annual Meeting of the Association
  for Computational Linguistics}, pages 8342--8360.

\bibitem[{Hayashi et~al.(2020)Hayashi, Budania, Wang, Ackerson, Neervannan, and
  Neubig}]{hayashi20tacl}
Hiroaki Hayashi, Prashant Budania, Peng Wang, Chris Ackerson, Raj Neervannan,
  and Graham Neubig. 2020.
\newblock \href {https://arxiv.org/abs/2011.07832} {Wikiasp: A dataset for
  multi-domain aspect-based summarization}.
\newblock \emph{Transactions of the Association for Computational Linguistics
  (TACL)}.

\bibitem[{Henry et~al.(2020{\natexlab{a}})Henry, Buchan, Filannino, Stubbs, and
  Uzuner}]{henry20202018}
Sam Henry, Kevin Buchan, Michele Filannino, Amber Stubbs, and Ozlem Uzuner.
  2020{\natexlab{a}}.
\newblock 2018 n2c2 shared task on adverse drug events and medication
  extraction in electronic health records.
\newblock \emph{Journal of the American Medical Informatics Association},
  27(1):3--12.

\bibitem[{Henry et~al.(2020{\natexlab{b}})Henry, Wang, Shen, and
  Uzuner}]{henry20202019}
Sam Henry, Yanshan Wang, Feichen Shen, and Ozlem Uzuner. 2020{\natexlab{b}}.
\newblock The 2019 national natural language processing (nlp) clinical
  challenges (n2c2)/open health nlp (ohnlp) shared task on clinical concept
  normalization for clinical records.
\newblock \emph{Journal of the American Medical Informatics Association},
  27(10):1529--1537.

\bibitem[{Hermann et~al.(2015)Hermann, Ko{\v{c}}isk{\`y}, Grefenstette,
  Espeholt, Kay, Suleyman, and Blunsom}]{hermannteaching}
KM~Hermann, T~Ko{\v{c}}isk{\`y}, E~Grefenstette, L~Espeholt, W~Kay, M~Suleyman,
  and P~Blunsom. 2015.
\newblock Teaching machines to read and comprehend.
\newblock \emph{Advances in Neural Information Processing Systems}, 28.

\bibitem[{Jackson et~al.(2015)Jackson, Shahsahebi, Wedlake, and
  Dubard}]{Jackson2015TimelinessOO}
Carlos~T. Jackson, Mohammad Shahsahebi, Tiffany Wedlake, and C~Annette Dubard.
  2015.
\newblock Timeliness of outpatient follow-up: An evidence-based approach for
  planning after hospital discharge.
\newblock \emph{Annals of Family Medicine}, 13 2:115--22.

\bibitem[{Jagannatha et~al.(2019)Jagannatha, Liu, Liu, and Yu}]{MADE}
Abhyuday Jagannatha, Feifan Liu, Weisong Liu, and Hong Yu. 2019.
\newblock Overview of the first natural language processing challenge for
  extracting medication, indication, and adverse drug events from electronic
  health record notes (made 1.0).
\newblock \emph{Drug Safety}, 42:99--111.

\bibitem[{Johnson et~al.(2016)Johnson, Pollard, Shen, wei H.~Lehman, Feng,
  Ghassemi, Moody, Szolovits, Celi, and Mark}]{mimiciii}
Alistair E.~W. Johnson, Tom~J. Pollard, Lu~Shen, Li~wei H.~Lehman, Mengling
  Feng, Mohammad~M. Ghassemi, Benjamin Moody, Peter Szolovits, Leo~Anthony
  Celi, and Roger~G. Mark. 2016.
\newblock Mimic-iii, a freely accessible critical care database.
\newblock In \emph{Scientific Data}.

\bibitem[{Johnson et~al.(2020)Johnson, Bulgarelli, and
  Pollard}]{johnson2020deidentification}
Alistair~EW Johnson, Lucas Bulgarelli, and Tom~J Pollard. 2020.
\newblock \href {https://doi.org/10.1145/3368555.3384455} {Deidentification of
  free-text medical records using pre-trained bidirectional transformers}.
\newblock In \emph{Proceedings of 2020 the Association for Computing Machinery
  (ACM) Conference on Health, Inference, and Learning (CHIL)}, pages 214--221.

\bibitem[{Kelly et~al.(2019)Kelly, Karthikesalingam, Suleyman, Corrado, and
  King}]{Kelly2019KeyCF}
Christopher~J. Kelly, Alan Karthikesalingam, Mustafa Suleyman, Greg Corrado,
  and Dominic King. 2019.
\newblock Key challenges for delivering clinical impact with artificial
  intelligence.
\newblock In \emph{BMC Medicine}.

\bibitem[{Kripalani et~al.(2007)Kripalani, LeFevre, Phillips, Williams,
  Basaviah, and Baker}]{Kripalani2007DeficitsIC}
Sunil Kripalani, Frank LeFevre, Christopher~O. Phillips, Mark~V. Williams,
  Preetha Basaviah, and David~W. Baker. 2007.
\newblock \href {https://doi.org/10.1001/jama.297.8.831} {{Deficits in
  Communication and Information Transfer Between Hospital-Based and Primary
  Care Physicians Implications for Patient Safety and Continuity of Care}}.
\newblock \emph{Journal of the American Medical Informatics Association
  (JAMIA)}, 297(8):831--841.

\bibitem[{Lee et~al.(2020)Lee, Yoon, Kim, Kim, Kim, So, and Kang}]{BioBERT}
Jinhyuk Lee, Wonjin Yoon, Sungdong Kim, Donghyeon Kim, Sunkyu Kim, Chan~Ho So,
  and Jaewoo Kang. 2020.
\newblock Biobert: A pre-trained biomedical language representation model for
  biomedical text mining.
\newblock \emph{Bioinformatics (Oxford, England)}, 36(4).

\bibitem[{Liang et~al.(2019)Liang, Tsou, and Poddar}]{liang-etal-2019-novel}
Jennifer Liang, Ching-Huei Tsou, and Ananya Poddar. 2019.
\newblock \href {https://doi.org/10.18653/v1/W19-1906} {A novel system for
  extractive clinical note summarization using {EHR} data}.
\newblock In \emph{Proceedings of the 2nd Clinical Natural Language Processing
  Workshop}, pages 46--54, Minneapolis, Minnesota, USA. Association for
  Computational Linguistics.

\bibitem[{Liu et~al.(2018{\natexlab{a}})Liu, Mojarad, Wang, Wang, Shen, Fu, and
  Liu}]{liu2018overview}
Sijia Liu, Majid~Rastegar Mojarad, Yanshan Wang, Liwei Wang, Feichen Shen,
  Sunyang Fu, and Hongfang Liu. 2018{\natexlab{a}}.
\newblock Overview of the biocreative/ohnlp 2018 family history extraction
  task.
\newblock In \emph{Proceedings of the BioCreative 2018 Workshop}, page 2018.

\bibitem[{Liu et~al.(2018{\natexlab{b}})Liu, Xu, Xie, and
  Xing}]{Semisupervised}
Xiangan Liu, Keyang Xu, Pengtao Xie, and Eric~P. Xing. 2018{\natexlab{b}}.
\newblock \href {http://arxiv.org/abs/1811.08040} {Unsupervised pseudo-labeling
  for extractive summarization on electronic health records}.
\newblock \emph{CoRR}, abs/1811.08040.

\bibitem[{Liu and Lapata(2019)}]{liu2019text}
Yang Liu and Mirella Lapata. 2019.
\newblock Text summarization with pretrained encoders.
\newblock In \emph{Proceedings of the 2019 Conference on Empirical Methods in
  Natural Language Processing and the 9th International Joint Conference on
  Natural Language Processing (EMNLP-IJCNLP)}, pages 3721--3731.

\bibitem[{McDermott et~al.(2020)McDermott, Hsu, Weng, Ghassemi, and
  Szolovits}]{mcdermott2020}
Matthew McDermott, Tzu Ming~Harry Hsu, Wei-Hung Weng, Marzyeh Ghassemi, and
  Peter Szolovits. 2020.
\newblock \href {https://arxiv.org/abs/2006.15229v1} {Chexpert++: Approximating
  the chexpert labeler for speed, differentiability, and probabilistic output}.
\newblock In \emph{Proceedings of Machine Learning Research}, volume 126. PMLR.

\bibitem[{Moore et~al.(2007)Moore, Mcginn, and Halm}]{Moore2007TyingUL}
Carlton Moore, Thomas Mcginn, and Ethan~A. Halm. 2007.
\newblock Tying up loose ends: Discharging patients with unresolved medical
  issues.
\newblock \emph{Archives of Internal Medicine}, 167 12:1305--11.

\bibitem[{Mullenbach et~al.(2021)Mullenbach, Pruksachatkun, Adler, Seale,
  Swartz, McKelvey, Yang, and Sontag}]{mullenbach2021}
James Mullenbach, Yada Pruksachatkun, Sean Adler, Jennifer Seale, Jordan
  Swartz, T.~Greg McKelvey, Yi~Yang, and David Sontag. 2021.
\newblock \href {https://doi.org/10.13026/6kfs-9v69} {Clip: A dataset for
  extracting action items for physicians from hospital discharge notes (version
  1.0.0)}.
\newblock \emph{PhysioNet}.

\bibitem[{Mulyar et~al.(2019)Mulyar, Schumacher, Rouhizadeh, and
  Dredze}]{BERTPhenotyping}
Andriy Mulyar, Elliot Schumacher, Masoud Rouhizadeh, and Mark Dredze. 2019.
\newblock Phenotyping of clinical notes with improved document classification
  models using contextualized neural language models.
\newblock \emph{ArXiv}, abs/1910.13664.

\bibitem[{Nallapati et~al.(2016)Nallapati, Zhou, dos Santos, Gulcehre, and
  Xiang}]{nallapati2016abstractive}
Ramesh Nallapati, Bowen Zhou, Cicero dos Santos, Caglar Gulcehre, and Bing
  Xiang. 2016.
\newblock Abstractive text summarization using sequence-to-sequence rnns and
  beyond.
\newblock In \emph{Proceedings of The 20th SIGNLL Conference on Computational
  Natural Language Learning}, pages 280--290.

\bibitem[{Pampari et~al.(2018)Pampari, Raghavan, Liang, and
  Peng}]{pampari2018emrqa}
Anusri Pampari, Preethi Raghavan, Jennifer Liang, and Jian Peng. 2018.
\newblock emrqa: A large corpus for question answering on electronic medical
  records.
\newblock In \emph{Proceedings of the 2018 Conference on Empirical Methods in
  Natural Language Processing}, pages 2357--2368.

\bibitem[{Ratner et~al.(2017)Ratner, Bach, Ehrenberg, Fries, Wu, and
  R{\'e}}]{ratner2017snorkel}
Alexander Ratner, Stephen~H Bach, Henry Ehrenberg, Jason Fries, Sen Wu, and
  Christopher R{\'e}. 2017.
\newblock Snorkel: Rapid training data creation with weak supervision.
\newblock In \emph{Proceedings of the VLDB Endowment. International Conference
  on Very Large Data Bases}, volume~11, page 269. NIH Public Access.

\bibitem[{Si et~al.(2020)Si, Wang, Wosik, Zhang, Dov, Wang, Henao, and
  Carin}]{si2020students}
Shijing Si, Rui Wang, Jedrek Wosik, Hao Zhang, David Dov, Guoyin Wang, Ricardo
  Henao, and Lawrence Carin. 2020.
\newblock \href {https://arxiv.org/abs/2006.11991v1} {Students need more
  attention: Bert-based attention model for small data with application to
  automatic patient message triage}.
\newblock In \emph{Proceedings of Machine Learning Research}, volume 126. PMLR.

\bibitem[{Singh et~al.(2013)Singh, Spitzmueller, Petersen, Sawhney, and
  Sittig}]{Singh2013InformationOA}
Hardeep Singh, Christiane Spitzmueller, Nancy~J. Petersen, Mona~K. Sawhney, and
  Dean~F. Sittig. 2013.
\newblock Information overload and missed test results in electronic health
  record-based settings.
\newblock \emph{JAMA Internal Medicine}, 173 8:702--4.

\bibitem[{Sinsky et~al.(2016)Sinsky, Colligan, Li, Prgomet, Reynolds, Goeders,
  Westbrook, Tutty, and Blike}]{Sinsky}
Christine Sinsky, Lacey Colligan, Ling Li, Mirela Prgomet, Sam Reynolds,
  Lindsey Goeders, Johanna Westbrook, Michael Tutty, and George Blike. 2016.
\newblock \href {https://doi.org/10.7326/M16-0961} {{Allocation of Physician
  Time in Ambulatory Practice: A Time and Motion Study in 4 Specialties}}.
\newblock \emph{Annals of Internal Medicine}, 165(11):753--760.

\bibitem[{Spencer et~al.(2019)Spencer, Rodgers, Salema, Campbell, and
  Avery}]{QualitativeInterview}
Rachel~A Spencer, Sarah Rodgers, Ndeshi Salema, Stephen~M Campbell, and
  Anthony~J Avery. 2019.
\newblock \href {https://bjgpopen.org/content/3/1/bjgpopen18X101625.abstract}
  {Processing discharge summaries in general practice: a qualitative interview
  study with gps and practice managers}.
\newblock \emph{British Journal of General Practice (BJGP)},
  https://doi.org/10.3399/bjgpopen18X101625.

\bibitem[{Suhr et~al.(2017)Suhr, Lewis, Yeh, and Artzi}]{suhr2017corpus}
Alane Suhr, Mike Lewis, James Yeh, and Yoav Artzi. 2017.
\newblock A corpus of natural language for visual reasoning.
\newblock In \emph{Proceedings of the 55th Annual Meeting of the Association
  for Computational Linguistics (Volume 2: Short Papers)}, pages 217--223.

\bibitem[{Sun et~al.(2013)Sun, Rumshisky, and Uzuner}]{i2b22012}
Weiyi Sun, Anna Rumshisky, and {\"O}zlem Uzuner. 2013.
\newblock Evaluating temporal relations in clinical text: 2012 i2b2 challenge.
\newblock \emph{Journal of the American Medical Informatics Association
  (JAMIA)}, 20 5:806--13.

\bibitem[{Tai-Seale et~al.(2017)Tai-Seale, Olson, J, AS, C, M, W, and
  HS}]{TaiEHRusage}
M~Tai-Seale, CW~Olson, Li~J, Chan AS, Morikawa C, Durbin M, Wang W, and Luft
  HS. 2017.
\newblock \href {https://doi.org/10.1377/hlthaff.2016.0811} {{Electronic Health
  Record Logs Indicate That Physicians Split Time Evenly Between Seeing
  Patients And Desktop Medicine.}}
\newblock \emph{Health Aff (Millwood)}, 1(35):655--662.

\bibitem[{Tenney et~al.(2019)Tenney, Das, and Pavlick}]{tenney2019bert}
Ian Tenney, Dipanjan Das, and Ellie Pavlick. 2019.
\newblock Bert rediscovers the classical nlp pipeline.
\newblock In \emph{Proceedings of the 57th Annual Meeting of the Association
  for Computational Linguistics}, pages 4593--4601.

\bibitem[{Uzuner et~al.(2012)Uzuner, Bodnari, Shen, Forbush, Pestian, and
  South}]{i2b22011}
{\"O}zlem Uzuner, Andreea Bodnari, Shuying Shen, Tyler Forbush, John Pestian,
  and Brett~R. South. 2012.
\newblock Evaluating the state of the art in coreference resolution for
  electronic medical records.
\newblock \emph{Journal of the American Medical Informatics Association
  (JAMIA)}, 19 5:786--91.

\bibitem[{Uzuner et~al.(2010)Uzuner, Solti, and Cadag}]{i2b22009}
{\"O}zlem Uzuner, Imre Solti, and Eithon Cadag. 2010.
\newblock Extracting medication information from clinical text.
\newblock \emph{Journal of the American Medical Informatics Association
  (JAMIA)}, 17 5:514--8.

\bibitem[{Uzuner et~al.(2011)Uzuner, South, Shen, and DuVall}]{i2b22010}
{\"O}zlem Uzuner, Brett~R. South, Shuying Shen, and Scott~L. DuVall. 2011.
\newblock 2010 i2b2/va challenge on concepts, assertions, and relations in
  clinical text.
\newblock \emph{Journal of the American Medical Informatics Association
  (JAMIA)}, 18 5:552--6.

\bibitem[{Wang et~al.(2019)Wang, Wang, Xiong, Yu, Guo, Chang, and
  Wang}]{wang2019self}
Hong Wang, Xin Wang, Wenhan Xiong, Mo~Yu, Xiaoxiao Guo, Shiyu Chang, and
  William~Yang Wang. 2019.
\newblock Self-supervised learning for contextualized extractive summarization.
\newblock In \emph{Proceedings of the 57th Annual Meeting of the Association
  for Computational Linguistics}, pages 2221--2227.

\bibitem[{Weis and Levy(2014)}]{weis2014copy}
Justin~M Weis and Paul~C Levy. 2014.
\newblock Copy, paste, and cloned notes in electronic health records.
\newblock \emph{Chest}, 145(3):632--638.

\bibitem[{Were et~al.(2010)Were, Gorbachev, Cadwallader, Kesterson, Li,
  Overhage, and Friedlin}]{NLPMartin}
Martin~C. Were, Sergey Gorbachev, Jason Cadwallader, Joe Kesterson, Xiaochun
  Li, J.~Marc Overhage, and Jeff Friedlin. 2010.
\newblock Natural language processing to extract follow-up provider information
  from hospital discharge summaries.
\newblock \emph{AMIA Annual Symposium Proceedings}, 2010:872--6.

\bibitem[{Wolf et~al.(2019)Wolf, Debut, Sanh, Chaumond, Delangue, Moi, Cistac,
  Rault, Louf, Funtowicz, Davison, Shleifer, von Platen, Ma, Jernite, Plu, Xu,
  Scao, Gugger, Drame, Lhoest, and Rush}]{Wolf2019HuggingFacesTS}
Thomas Wolf, Lysandre Debut, Victor Sanh, Julien Chaumond, Clement Delangue,
  Anthony Moi, Pierric Cistac, Tim Rault, Rémi Louf, Morgan Funtowicz, Joe
  Davison, Sam Shleifer, Patrick von Platen, Clara Ma, Yacine Jernite, Julien
  Plu, Canwen Xu, Teven~Le Scao, Sylvain Gugger, Mariama Drame, Quentin Lhoest,
  and Alexander~M. Rush. 2019.
\newblock Huggingface's transformers: State-of-the-art natural language
  processing.
\newblock \emph{ArXiv}, abs/1910.03771.

\bibitem[{Zhang et~al.(2020)Zhang, Lu, Abdalla, McDermott, and
  Ghassemi}]{zhang2020hurtful}
Haoran Zhang, Amy~X Lu, Mohamed Abdalla, Matthew McDermott, and Marzyeh
  Ghassemi. 2020.
\newblock \href {https://doi.org/10.1145/3368555.3384448} {Hurtful words:
  Quantifying biases in clinical contextual word embeddings}.
\newblock In \emph{Proceedings of the ACM Conference on Health, Inference, and
  Learning}, pages 110--120.

\bibitem[{Zhang et~al.(2019)Zhang, Chen, Yang, Lin, and Lu}]{biowordvec1}
Yijia Zhang, Qingyu Chen, Zhihao Yang, Hongfei Lin, and Zhiyong Lu. 2019.
\newblock Biowordvec, improving biomedical word embeddings with subword
  information and mesh.
\newblock In \emph{Scientific Data}.

\end{thebibliography}
\clearpage
\appendix

\section{Additional data processing details}

When a focus sentence is near the start or end of a document, we use special \texttt{<DOC\_START>} and \texttt{<DOC\_END>} tokens in place of sentences, as needed when the limits of the document are reached. Because BERT takes a maximum length of 512 tokens, and due to occasional long sentences in MIMIC, when including context we may have to truncate our input, which occurs for just under 1\% of sentences. To do this, we first remove the shorter of the two outermost context sentences, then remove other context sentences as needed, alternating sides and moving inward. Finally, the tokenized input along with position and segment embeddings is fed into the transformer layers to obtain contextualized representations for each token. 

\section{Additional annotation details}

We built an internal annotation tool, which allowed annotators to select and label arbitrary character-level spans of text within the document.
These character-level spans were later converted into the sentence annotations.

Since MIMIC-III is an anonymized dataset, entities such as names, dates, phone numbers, hospital names, and others censored and replaced with a templated substitute. We apply a surrogate generation process to fill in synthetic entities in place of these templates, to make reading and annotating notes easier. These surrogates are also used at prediction time. Due to space constraints, the full guidelines are provided on our public GitHub repository \footnote{\url{https://github.com/asappresearch/clip}}.

\subsection{Annotation refinement}
After collecting initial annotations, we met with the annotators in multiple sessions to reconcile differences in their annotations. We adjusted the annotation guidelines slightly to reduce ambiguity and improve labeling consistency. Using an initial set of examples that were annotated by multiple experts, we identified examples where labels disagreed across annotators. In the reconciliation process, we discussed those disagreements as a group to determine whether (a) one annotator misapplied or forgot to apply an annotation guideline, or (b) the proper annotation was ambiguous given the guidelines at that time. In the case of ambiguous guidelines, we would then add a new rule or example to the guidelines. The sources of disagreement were commonly in the ``Other'' category, which encompasses a long tail of information - we capture these in the guidelines with a non-comprehensive set of examples demonstrating both labeled and unlabeled cases.

The \textsc{Patient Instructions} label originally instructed annotators to choose only those instructions that are unique to that patient, and exclude general guidelines such as ``Call your doctor if you experience a fever.'' However, we observed this was too ambiguous in practice, so we chose to automatically label any sentence in document sections ``Followup instructions'' and ``Discharge instructions'' as the \textsc{Patient Instructions} label, using regular expressions to identify common section headers in the MIMIC-III discharge notes. Our annotators provided additional manual annotations, so not all examples of this type come from these rule-derived annotations; any annotations with the ``Patient instructions'' label which appear outside of the ``Followup instructions'' and ``Discharge instructions'' sections were manually annotated. 

Finally, two of the original annotators revised all existing annotations, to catch mistakes and adjust to the refined guidelines. 

\section{CLIP dataset phenomena details}

In \autoref{tab:clip_phenomena_details} we provide a breakdown of high-level phenomena in the CLIP dataset by label type. We sampled sentences randomly, ensuring each label type had at least 20 examples.

\begin{table*}[t]
\centering
\begin{tabular}{@{}lrrrrrr@{}}
\toprule
\textbf{Label} & \multicolumn{1}{l}{\textbf{N}} & \multicolumn{1}{l}{\textbf{\# OOV}} & \multicolumn{1}{l}{\textbf{Quantities}} & \multicolumn{1}{l}{\textbf{Temporal}} & \multicolumn{1}{l}{\textbf{Conditional}} & \multicolumn{1}{l}{\textbf{Imperative}} \\ \midrule
Imaging & 20 & 1.70 & 0.40 & 0.70 & 0.05 & 0.45 \\
Appointment & 47 & 2.02 & 0.19 & 0.62 & 0.11 & 0.32 \\
Medication & 25 & 2.72 & 0.64 & 0.52 & 0.04 & 0.16 \\
Procedure & 24 & 1.71 & 0.21 & 0.50 & 0.17 & 0.33 \\
Lab & 23 & 1.87 & 0.26 & 0.44 & 0.09 & 0.39 \\
Patient & 78 & 1.63 & 0.18 & 0.50 & 0.14 & 0.58 \\
Other & 21 & 2.33 & 0.00 & 0.29 & 0.10 & 0.24 \\
All & 160 & 2.00 & 0.25 & 0.48 & 0.11 & 0.38 \\ \bottomrule
\end{tabular}
\caption{Observed phenomena for a random selection of each label type. \# OOV is an average across sentences, while the other measures are fractions.}
\label{tab:clip_phenomena_details}
\end{table*}

\clearpage

\end{document}